\documentclass[letterpaper, 10 pt, conference]{ieeeconf}  

\IEEEoverridecommandlockouts                              

\usepackage{graphicx}
\usepackage{booktabs} 
\usepackage{multirow} 
\usepackage{float}
\usepackage{siunitx}
\usepackage{amsmath, amsfonts}
\usepackage[normalem]{ulem}
\usepackage{multirow}
\usepackage{bm}
\let\labelindent\relax
\usepackage{enumitem}
\usepackage{longtable}
\usepackage[table]{xcolor}
\usepackage{subcaption}
\usepackage{url}

\title{\LARGE \bf DexSkills: Skill Segmentation Using Haptic Data for Learning Autonomous Long-Horizon Robotic Manipulation Tasks}

\author{
    Xiaofeng Mao$^{1,\dag}$ and Gabriele Giudici$^{2,\dag}$,\\
    Claudio Coppola$^{3}$, Kaspar Althoefer$^{2}$, Ildar Farkhatdinov$^{2}$, Zhibin Li$^{4}$, Lorenzo Jamone$^{2}$
\thanks{Xiaofeng Mao and Gabriele Giudici contributed equally to this work and share first authorship.}
\thanks{Work not related to Claudio Coppola's position in Amazon.}
\thanks{$^{1}$University of Edinburgh, email: xiaofeng.mao@ed.ac.uk}
\thanks{$^{2}$The authors are with ARQ  (the Centre for Advanced Robotics @ Queen
Mary), School of Engineering and Materials Science, Queen Mary University of London, London, E14NS, UK (emails: \{g.giudici,k.althoefer,i.farkhatdinov,l.jamone\}@qmul.ac.uk).}%
\thanks{$^{3}$Amazon Transportation Services, email: claudcop@amazon.co.uk}    
\thanks{$^{4}$University College London, email: alex.li@ucl.ac.uk}
\thanks {The study is funded by the UKRI EPSRC grants EP/R02572X/1 (NCNR), EP/V035304/1 (q-Arena) and Queen Mary University of London Ph.D. scholarship to G. Giudici.}
}

\begin{document}

\maketitle
\thispagestyle{empty}
\pagestyle{empty}

\begin{abstract}

Effective execution of long-horizon tasks with dexterous robotic hands remains a significant challenge in real-world problems. While learning from human demonstrations have shown encouraging results, they require extensive data collection for training. Hence, decomposing long-horizon tasks into reusable primitive skills is a more efficient approach. To achieve so, we developed DexSkills, a novel supervised learning framework that addresses long-horizon dexterous manipulation tasks using primitive skills. DexSkills is trained to recognize and replicate a select set of skills using human demonstration data, which can then segment a demonstrated long-horizon dexterous manipulation task into a sequence of primitive skills to achieve one-shot execution by the robot directly. Significantly, DexSkills operates solely on proprioceptive and tactile data, i.e., haptic data. Our real-world robotic experiments show that DexSkills can accurately segment skills, thereby enabling autonomous robot execution of a diverse range of tasks.

\end{abstract}

\section{INTRODUCTION}
Humans show remarkable dexterity and adaptability in performing manipulation tasks across various environments, this is attributed to the inherent capabilities of the human hand.
However, enabling robotic dexterous manipulation with multi-fingered hands remains challenging due to the high-dimensional action space and occlusion during manipulation.
Recent research efforts have focused on addressing this challenge through either model-based control methods~\cite{yao2023exploiting, gao2023real, khadivar2023adaptive} or learning-based approaches~\cite{petrenko2023dexpbt, handa2023dextreme, sun2023dexdlo}.
However, these methods rely on complex mathematical models, require customized rewards designed by humans, demand extensive training time, and necessitate large amounts of demonstration data~\cite{yu2022dexterous}.
Consequently, the training processes become inefficient and labor-intensive, hindering robots from achieving dexterous manipulation.
%

Multi-modal information is crucial in enabling robot dexterous manipulation.
In routine tasks like pick-and-place operations, visual cues provide information on the position and shape of the object, aiding the robot during its execution.
However, the availability of accurate visual information may be compromised due to factors such as poor lighting conditions or occlusions. 
In addition, certain aspects of objects and tasks, especially in contact-rich manipulation tasks, are better encoded by tactile and force sensing \cite{luo2017robotic,li2020review}.
Tactile feedback offers valuable insights into the contact dynamics between the robotic fingers and the manipulated object, enhancing the potential for accurate and stable manipulation. Hence, numerous studies are currently investigating the viability of leveraging tactile information as an alternative or complementary source \cite{jiang2023robotic, yin2023rotating, lee2024dextouch, pai2023tactofind, hu2023dexterous}. 

\begin{figure}[t]
\centering
\includegraphics[width=1.0\linewidth,keepaspectratio]{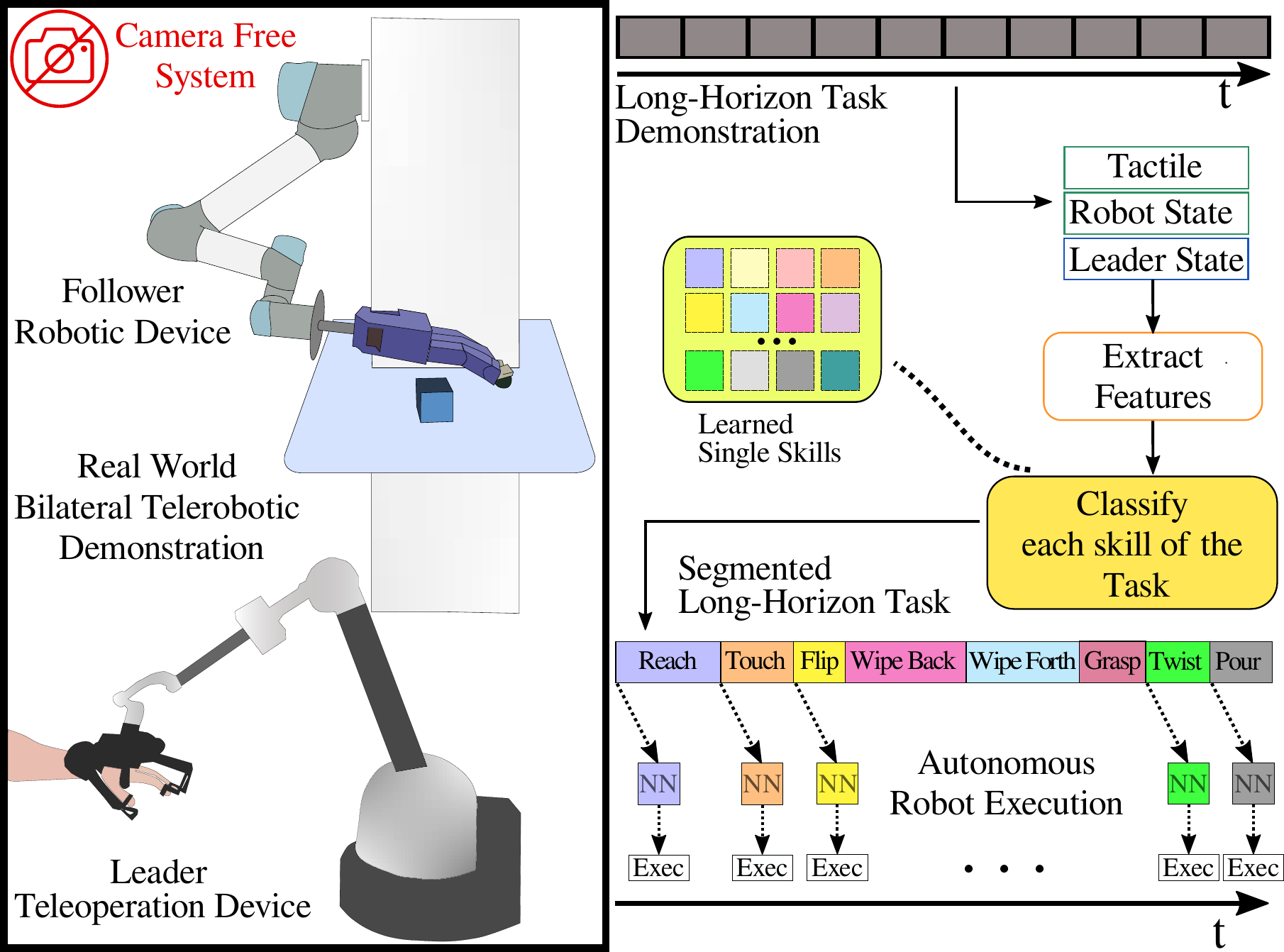}
\caption{
Overview of the proposed long-horizon task segmentation approach. Individual skills are segmented and classified at each temporal window of the demonstration. The demonstrations are collected via the teleoperation system (left) developed in \cite{giudici2023feeling}.
}
\label{Fig:Figure_1}
\end{figure}

Learning from human demonstrations makes it possible to learn from real-world human demonstrations, avoiding the need for intricate dynamic models, well defined reward functions, and reliable simulated environments \cite{elguea2023review}.
On the other hand, these approaches often demand a considerable amount of data and human effort \cite{correia2023survey}.
Moreover, for long-horizon tasks, these methods are prone to failure due to the accumulation of compounding errors and necessitate model retraining for each new demonstrated task, even if many task phases are in common between tasks \cite{luo2024multi, shi2023waypoint}.

Therefore, we propose DexSkills, a system that learns primitive skills from human demonstrations, and uses such knowledge to segment a human-demonstrated long-horizon task in a sequence of skills that the robot can execute autonomously. This requires less data collection efforts with respect to learning each possible long-horizon task from human demonstration. Notably, DexSkills relies only on proprioceptive and tactile data (i.e. haptic data). In particular, the contributions of this work are: 




%
%
%

\begin{enumerate}

\item We introduce a novel supervised representation learning framework where the latent features from haptic data are jointly trained via an auto-regressive autoencoder and a label decoder. Our qualitative result proves that by capturing the latent dynamics of primitive skills, the performance of segmenting unseen long-horizon tasks into skill sequences is significantly improved;




\item We propose a set of 20 core primitive skills specifically designed to address robot dexterous manipulation tasks. We trained the autonomous robot controller for each skill using Multi-Layer Perceptron (MLP) and demonstrated that by recombining these trained core primitive skills, the robot is able to compose and achieve long-horizon dexterous tasks effectively and successfully;



\item To facilitate future research, we open-sourced our code\footnote{\label{footnote:Dexskills}\url{https://github.com/ARQ-CRISP/DexSkills}} and provided a labeled dataset comprising 20 primitive skills and 20 long-horizon tasks, which were obtained from human teleoperated demonstrations.



\end{enumerate}

DexSkills has developed a primary set of reusable primitive skills trained from haptic human demonstrations, which can be reused and re-combined to represent various compositions of long-horizon tasks. DexSkills achieves remarkable segmentation accuracy of $91\%$ for unseen tasks and enables autonomous robot execution of diverse tasks using only proprioceptive and tactile data.

\section{Related Works}

\textbf{Imitation learning based robot dexterous manipulation:}
Robot dexterous manipulation has recently gained increasing attention~\cite{billard2019trends}. 
Imitation learning (IL) is a well-established method that accelerates the skill learning process in robots by leveraging human experience~\cite{ravichandar2020recent}. 
In~\cite{kumar2016learning}, human demonstrations are captured within a virtual environment, employing learning-based methods for feedback control in executing non-prehensile manipulations
The research~\cite{radosavovic2021state} proposes to train an inverse dynamics model to predict actions for state-only demonstration~\cite{radosavovic2021state}. 
In~\cite{arunachalam2023dexterous}, an RGB camera is used to observe a human operator teleoperate a robot and train the policy with the imitation learning method. 

\textbf{Long-horizon imitation learning:}
For long-horizon manipulation tasks, traditional IL methods might not be sufficient and could fail due to the accumulation of compounding errors. Breaking down the long-horizon manipulation task into multiple sub-tasks is a widely adopted strategy for addressing this challenge.
By decomposing the IL for manipulation task into visual servoing and behaviour replay, \cite{valassakis2022demonstrate} has effectively facilitated one-shot imitation learning for real-world daily tasks. 
Additionally, the strategy of automatic waypoint extraction to segment the entire trajectory significantly reduces the BC horizon, effectively mitigating the problem of error accumulation~\cite{shi2023waypoint}.
Hierarchical imitation learning serves as a prominent strategy to solve long-horizon manipulation tasks. By training a high-level policy for skill selection and a low-level policy for precise motor control, robots are capable of handling a wide range of dexterous manipulation tasks, including activities like cable routing~\cite{luo2024multi}, drink pouring~\cite{zhang2021explainable} and bi-manual cooperation task~\cite{xie2020deep, zhao2023learning}. 

\textbf{Dexterous manipulation with tactile sensing:}
Recently, some works have investigated using tactile sensing to mitigate occlusion during manipulation tasks with robot dexterous hand. 
By overlaying the binary force sensor to the one side of the whole robot hand, the work~\cite{yin2023rotating} has successfully facilitated the object in-hand rotation task. 
With the same robot hand setup, the study~\cite{lee2024dextouch} demonstrated the capability of employing tactile sensors to search for and locate target objects, subsequently manipulating these objects to perform daily tasks. 
Similarly, \cite{guzey2023dexterity} use the tactile sensor Uskin~\cite{tomo2018new} and proposes to pre-train a tactile encoder to extract features from high-dimensional tactile information, which enables the robot to perform tasks such as book opening, bowl and cup unstacking. 
The work~\cite{pai2023tactofind} investigates using sparse tactile feedback to localize, identify and grasp novel objects without any visual feedback. 

\textbf{Primitive Skills:}
Daily life manipulation tasks often comprise a variety of primitive skills. Identifying the sequence in which these skills combine for any long-horizon task is crucial for facilitating their reuse, thereby broadening the scope of tasks that can be accomplished. Retrieval skills from long sequence demonstration can be achieved by using supervised~\cite{coppola2022master, pirk2020modeling} and unsupervised method~\cite{krishnan2017transition, meli2021unsupervised, zhu2022bottom, wan2023lotus, park2023controllability}.
For the reuse of the primitive skills, \cite{bharadhwaj2023roboagent} proposes to train a universal agent with 12 unique skills that are capable of multi-task manipulation by using semantic augmentation and action representation. 
The research by~\cite{triantafyllidis2023hybrid} proposes a hybrid hierarchical learning framework named ROMAN, which combines behavioural cloning (BC), reinforcement learning (RL), and IL. A central manipulation network is used in this work to produce the appropriate sequential actions for various sub-skills networks, aiming to solve complex long-horizon manipulation tasks. 
The work~\cite{chen2023sequential} proposes to chain multiple dexterous policies for achieving long-horizon dexterous manipulation tasks by defining a feasibility function. Different from previous works, we introduce primitive skills specifically designed for dexterous manipulation tasks, aimed at achieving unseen long-horizon tasks by predicting the sequence of primitive skills combination. Inspired by the work~\cite{coppola2022master}, we introduce a specialized set of features designed for telerobot dexterous manipulation, aiming to substantially enhance classifier performance through the integration of auto-regressive autoencoder (AE).  Moreover, we have developed and trained a distinct skill policy for each primitive, employing the predicted skill sequence to successfully execute long-horizon manipulation tasks. 








\section{METHODS}
In this study, our objective is to explore methods for accomplishing imitation learning for long-horizon dexterous manipulation tasks. The framework for learning long-horizon dexterous manipulation tasks is illustrated in Fig~\ref{Fig:Figure_1}. We start by proposing primitive skills for tasks involving the use of a robotic arm and dexterous hand. Our research focuses on learning these primitive skills with haptic data, segmenting the skill sequence from unseen long-horizon dexterous manipulation tasks, and executing the task by sequentially performing the identified skill segments. 

\subsection{Learning Features}\label{sec:learning_features}

For the robotic setup, which includes a robotic arm, a dexterous hand, and tactile sensors, we propose a set of features designed to differentiate primitive skills. These features are summarized as follows:
\subsubsection*{End Effector (EE) Information}
For the features related to the EE,  we utilize both position and velocity information to accurately capture the movements. Additionally, to provide a clearer representation of the movement of EE direction, we incorporate the EE direction as a feature. This direction is defined with a value into the set $[-1, 0, 1]$. Specifically, when the velocity exceeds $0.02 m/s$, the EE direction is designated as 1 to indicate forward movement. Conversely, if the velocity is less than $-0.02 m/s$, the direction is set to -1, indicating backward movement. In scenarios where the velocity of the EE falls within the threshold, the direction is set to 0, denoting a stationary state. These features yield a detailed understanding of the movement of the EE throughout the execution of tasks.
\begin{itemize}
    \item EE Pose: $\begin{bmatrix} x,y,z,\alpha,\beta,\gamma \end{bmatrix}$
    \item EE Velocity: $\begin{bmatrix} \dot{x},\dot{y},\dot{z},\dot{\alpha},\dot{\beta},\dot{\gamma} \end{bmatrix}$
    \item EE Direction: $[-1,0,1]$
\end{itemize}

\subsubsection*{Allegro Hand (AH) Information}
The features of the Allegro Hand (AH) include the AH joint state, fingertip positions, velocities, as well as position and velocity covariance. To effectively capture the movement correlations between each finger, the $Log$ of the upper triangle of the covariance matrix is utilized. This approach is chosen for its efficiency in representing finger movements during manipulation tasks.
\begin{itemize}
    \item AH Joint State: $[16 \times 1]$

    \item Fingerfitp Position: $[x,y,z]$
    \item Fingertips Velocity: $[\dot{x},\dot{y},\dot{z}]$
    \item Fingertip Position Covariance: [$triu(Log(C_p))$]
    \item Fingertip Velocity Covariance: [$triu(Log(C_v))$]
\end{itemize}

\subsubsection*{Tactile Information}
A fingertip with a custom-made magnetic sensor, presented in \cite{giudici2023feeling}, is attached to each finger of the AH. To quantify the intensity of force applied on each fingertip, we selected the norm of the tactile force as the feature. This allows us to accurately capture and analyze the dynamic interactions between the robot fingertips and manipulated objects. Additionally, a feature indicating the contact status is selected to reflect the interaction scenario between the AH and the object. This feature essentially distinguishes whether there is direct contact between the fingertip and the object. Incorporating this feature also aids in reducing noise, thereby enhancing the reliability and clarity of the sensory data used for task analysis and execution.

\begin{itemize}
    \item Tactile Norm $(Fx,Fy,Fz) \times 4$
    \item Contact Status: $[(0,1) \times 4]$
\end{itemize}

\begin{figure}[t]
\centering
\includegraphics[width=1.0\linewidth,keepaspectratio]{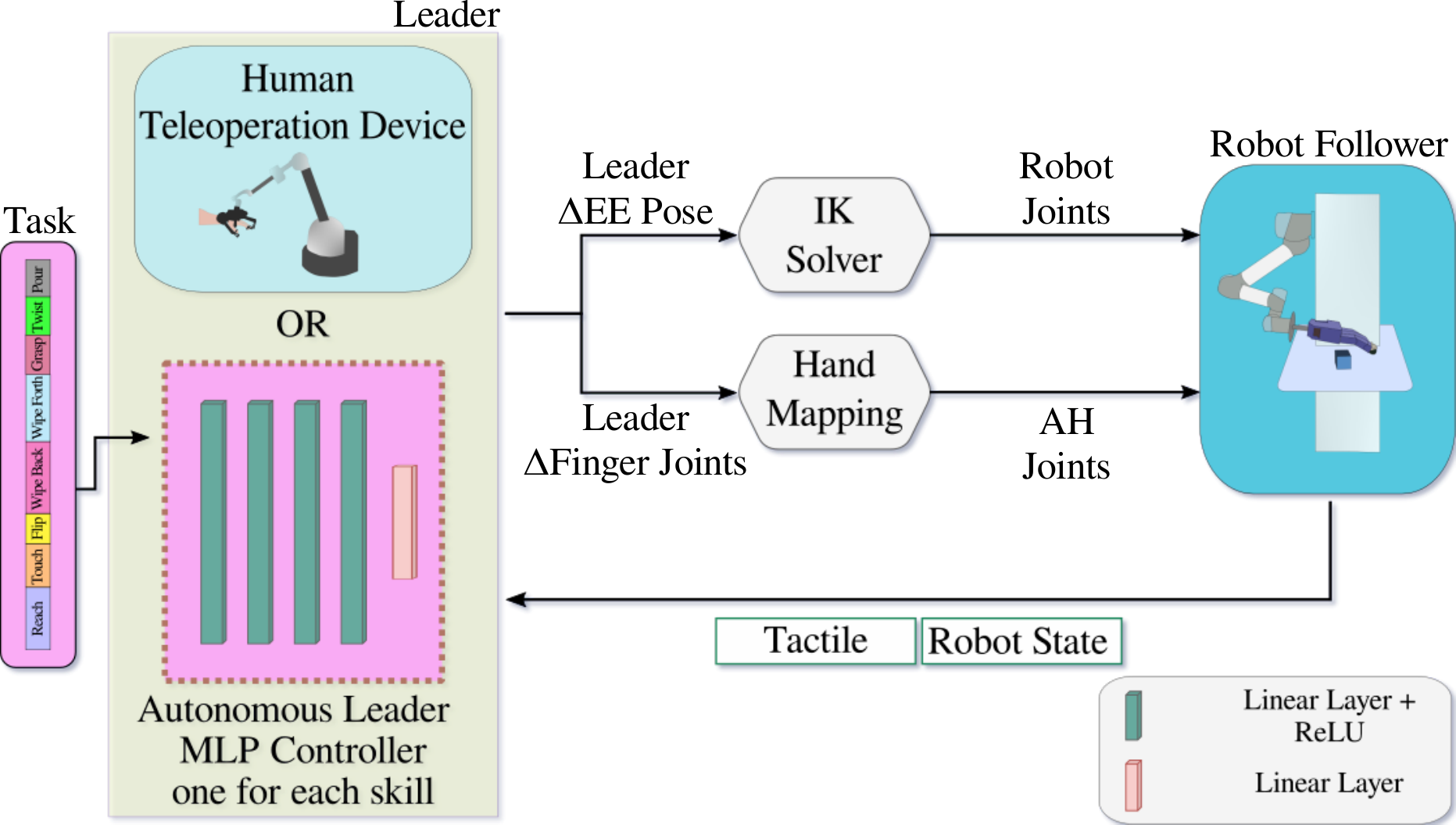}
\caption{
The leader agent generates motor control commands for the end effector pose and finger joints of the hand. The follower robot executes corresponding actions based on these commands. During teleoperation, the follower robot provides haptic feedback. 
When operating the robot autonomously, we control the robot using a distinct MLP trained on the proprioceptive and tactile data (i.e. haptic data) of each separate skill.
}
\label{Fig:fig3Controlarch}
\end{figure}

\begin{figure}[t]
\centering
\includegraphics[width=1.0\linewidth,keepaspectratio]{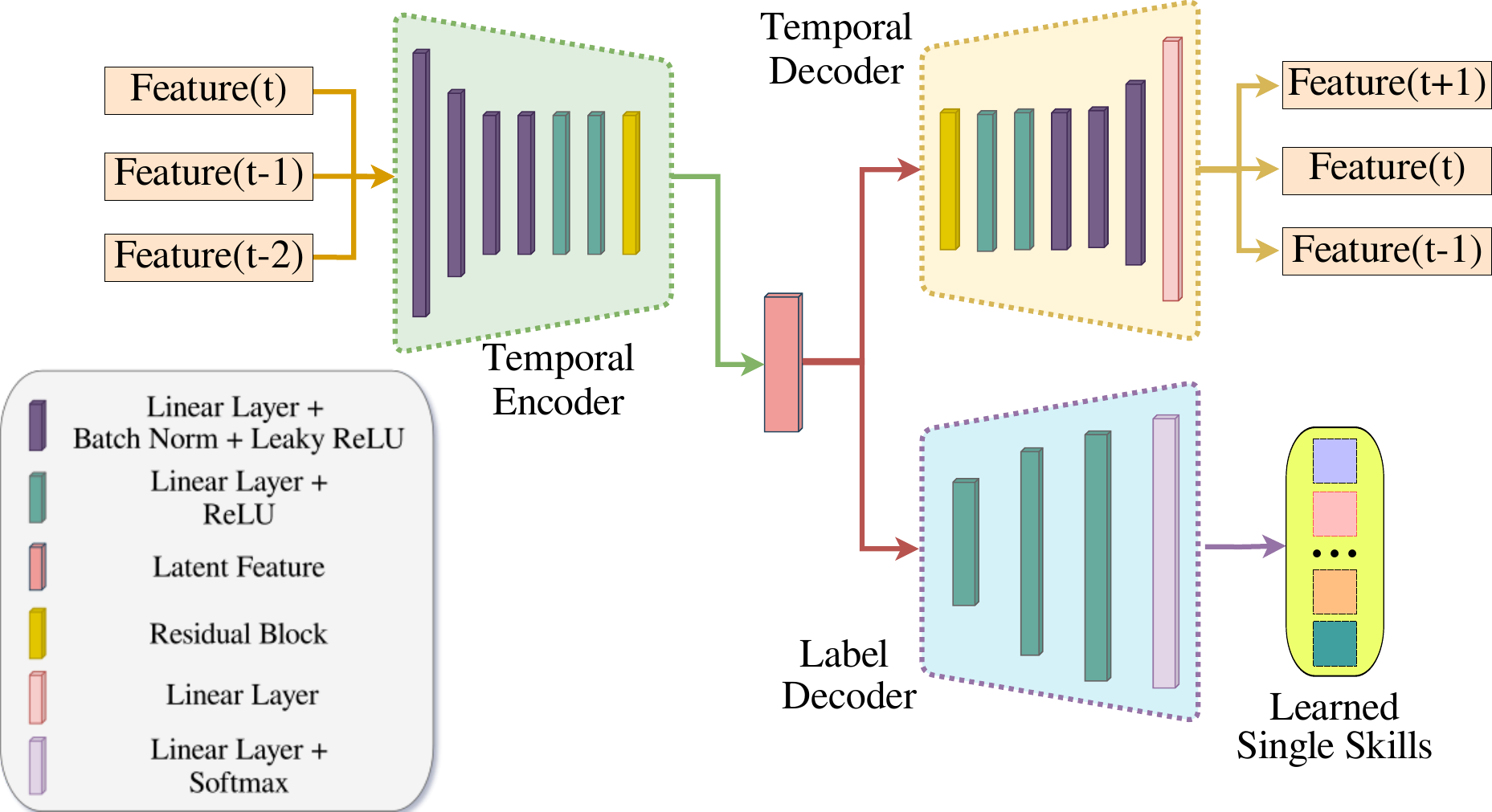}
\caption{
The architecture of our Neural Network for supervised representation learning incorporates an auto-regressive autoencoder and a label decoder.
This network processes time-series feature data as input, with the encoder transforming these features into a latent space. The temporal decoder reconstructs the features along with their predictions, whereas the label decoder extracts labels from the latent vectors. 
The label decoder is jointly trained with the autoencoder generating latent features that improve the segmentation performance. 
}
\label{Fig:fig2NNmodel}
\vspace{-5mm}
\end{figure}



\subsection{Primitive Skills Learning}

Long-horizon dexterous manipulation tasks often demand extensive demonstrations and face challenges related to accumulating errors and limited generalization capabilities.
When learning long-horizon dexterous manipulation skills, humans adeptly analyze and sequence primitive skills. 
Motivated by this human approach,  our study introduces a methodology for separately acquiring primitive skills and subsequently predicting the integration of these skills for unseen tasks. This strategy is aimed at equipping robots with the capability to perform long-horizon dexterous manipulation tasks. The specific primitive skills we proposed for robotic manipulation using dexterous hands are detailed in Table~\ref{tab:skills}.

The process of collecting primitive skills datasets and training the autonomous skill policy is shown in Fig.~\ref{Fig:fig3Controlarch}. The single manipulation policy operates with the assumption that the robot already knows the location of the object. 
For each primitive skill, an MLP neural network is utilized to learn the state-action pair from human demonstrations using the BC method. Given state information  $s$, the goal of the BC is to train a policy $\pi(\theta_p)$ that can learn the behaviour of the human demonstrator  by predicting the action $a$:

\begin{equation}
\theta_p=\arg \max _{\theta_p} \underset{\mathbf{s}, \mathbf{a} \sim \mathcal{D}}{\mathbb{E}} \left[L(\mathbf{a} \mid \mathbf{s})\right]
\end{equation}
where the observation $f$ is the input to the network, which includes the robot EE state, AH joint state, and filtered tactile information as well as the contact indicator. $L(\cdot)$ represents the loss function. In this work, the mean squared error (MSE) is used to calculate the loss between the predicted action and the recorded human action during the demonstration. 
The network output $a = [\Delta EE, \Delta Fingertip]$ captures the movement of both EE and AH fingertip, reflecting the actions executed by humans during the teleoperation of the robot system. This approach enables the policy $\pi(\theta_p)$ to directly learn from human input, controlling the robot through the same mapping to bypass the complexities of joint-space control and enhance safety. 


\subsection{Supervised Representation Learning}
The proposed framework is shown in Fig.~\ref{Fig:fig2NNmodel}. This segmentation framework consists of two parts.  Firstly, a temporal auto-regressive AE is utilized to encode temporal features into a latent space and then reconstruct the features for the subsequent time step. Secondly, a label decoder extracts the task label from the encoded latent representation. The AE receives an input sequence represented by $[\mathbf{f}_{t-2}, \mathbf{f}_{t-1}, \mathbf{f}_{t}]$, while the output generated by the encoder is $[\mathbf{f}_{t-1}, \mathbf{f}_{t}, \mathbf{f}_{t+1}]$, where $f$ denotes the features outlined in Section~\ref{sec:learning_features}. The interval between each timestep in this process is set to 0.1 seconds. Furthermore, the representation of skill labels employs a one-hot encoding scheme.

The choice of an auto-regressive AE is strategic for encapsulating the temporal dynamics of robot behaviour, crucial for capturing the intricate nature of robot actions where the sequence of actions is significant. This setup is designed to capture the nuances of human behaviour through temporal data analysis. Additionally, the AE facilitates a deeper comprehension of the input data, which in turn, augments the accuracy of task label predictions by understanding the underlying patterns and characteristics of human behavior encoded in the temporal information. During the training, we train the AE together with the label decoder. The loss function is defined as:
\begin{equation}
\text{Loss} = \text{MSE}(f_{\text{true}}, f_{\text{pred}}) + \text{CrossEntropy}(l_{\text{true}}, l_{\text{pred}})
\end{equation}
where the loss $\text{MSE}(f_{\text{true}}, f_{\text{pred}})$ calculates the MSE between the true temporal features $f_{\text{true}}$ and the predicted temporal features $f_{\text{pred}}$.
The loss function $\text{CrossEntropy}(l_{\text{true}}, l_{\text{pred}})$ is specifically selected to calculate the discrepancy between the actual probabilities $l_{\text{true}}$ and the predicted probabilities $l_{\text{pred}}$ for each class. The combination of these two loss functions encourages the model to learn rich and informative information that helps in both accurately predicting temporal features and correctly identifying skill labels.


\subsection{Unknown long-horizon manipulation skills learning}\label{sec::LH_method}
Our objective is to facilitate one-shot imitation learning for long-horizon dexterous manipulation tasks that were previously unseen, utilizing a set of predefined primitive skills. Segmenting these tasks into simpler skills makes it more manageable for the algorithm to understand and master the task, reducing the overall complexity. For each demonstration, robot data are gathered, and features are extracted from this collected data. 
The trained feature encoder $\mathcal{E}$ and the label decoder $\mathcal{L}$ are used to predict the sequence of primitive skill for executing dexterous long-horizon manipulation tasks:
\begin{equation}
\mathbf{P_t} = \mathcal{L}(\mathcal{E}(\mathbf{f}_{t-2}, \mathbf{f}_{t-1}, \mathbf{f}_t; \theta_\mathcal{E}); \theta_\mathcal{L}) ,
\end{equation}
where $\mathbf{P_t}$ is the predicted probability distribution of primitive skill at timestep $t$, $\theta_\mathcal{E}$ and $\theta_\mathcal{L}$ are the parameters of the feature encoder and label decoder respectively. To enhance the accuracy of these predictions and reduce errors, a median filter is applied for smoothing. Upon establishing a sequence for the primitive skills required for these dexterous long-horizon tasks, the trained primitive skills are executed in sequence to complete the tasks.





\section{Experimental Setting}

\subsection{Teleoperation Setup}

Depicted in Fig.~\ref{Fig:Figure_1}, the teleoperation setup used is similar to the one proposed by the authors in a previous work \cite{giudici2023feeling}. In summary, the setup consists of two leader devices: Virtuose 6D and HGlove, used to respectively control the robotic arm UR5 and Allegro dexterous Hand mounted with custom-made magnetic sensors on the follower side. The mapping and control structure is the same as in the previous study, as referenced and summarized in Fig.~\ref{Fig:fig3Controlarch}. In this study, the wrist rotation of the robotic arm is enabled in order to achieve a wider range of skills.


For the data collection phase, all the information used to control the robot and to extract the training features proposed in Sec.\ref{sec:learning_features} was recorded and made available in a dataset on the project's repository\footnotemark[\value{footnote}]. 




    

\subsection{The Primitive Skills}

In this work, amongst the various possible robotics manipulation skills, 20 primitive skills have been identified and are listed with a unique colour associated in Table \ref{tab:skills}. A combination of more than three skills defines a Long-Horizon task. Although all skills are shown independently, some of them require a pre-action in the real world to be performed. The task of planning long-horizon tasks is left to the demonstrator. 
The objective of this work is to provide a paradigm for segmentation and learning of skills where the feasibility of a task is defined by the demonstration sequences. For this reason, the assumption is made that each task is demonstrable and repeatable in a real-world scenario.

\begin{table}[t]
\centering
\caption{List of primitive skills}
\begin{tabular}{|c|c|c|}
\hline
\multirow{2}{*}{\textbf{Group}} & \multicolumn{2}{c|}{\textbf{Skills}} \\ \cline{2-3} 
                                 & \textbf{Pre}   & \textbf{Skill}   \\ \hline

\multirow{2}{*}{\shortstack{No Touch}}               & \cellcolor{white}-               & \cellcolor{blue!25}1. Reach\\ \cline{2-3} 
                                                         & \cellcolor{white}-               &\cellcolor{yellow!25}2. Setup Position\\ \hline                                 
\multirow{20}{*}{Touch}                                   &\cellcolor{red!25}3. PreTouch        &\cellcolor{orange!50}4. Touch                          \\ \cline{2-3} 
                                                         & \cellcolor{white}-               &\cellcolor{yellow!75}5. Flip\\ \cline{2-3} 
                                                         & \cellcolor{red!25}PreTouch&\cellcolor{cyan!25}6. Wipe Forth\\ \cline{2-3} 
                                                         & \cellcolor{red!25}PreTouch&\cellcolor{magenta!50}7. Wipe Back\\ \cline{2-3} 
                                                         &\cellcolor{violet!25}8. PreGrasp&\cellcolor{purple!50}9. Grasp\\ \cline{2-3}
                                                         & \cellcolor{white}-               &\cellcolor{teal!50}10. Lift with Grasp\\ \cline{2-3}                        
                                                         & \cellcolor{white}-               &\cellcolor{brown!50}11. Transport Forward\\ \cline{2-3} 
                                                         & \cellcolor{white}-               &\cellcolor{lime!75}12. Place                     
                                                         \\ \cline{2-3} 
                                                         &\cellcolor{olive!50}13. PreRotate       &\cellcolor{orange!75}14. Rotate                        
                                                         \\ \cline{2-3} 
                                                         & \cellcolor{white}-               &\cellcolor{cyan!75}15. Shake Up\\ \cline{2-3} 
                                                         & \cellcolor{white}-               &\cellcolor{magenta!75}16. Shake Down\\ \cline{2-3} 
                                                         & \cellcolor{white}-               &\cellcolor{green!75}17. Twist\\ \cline{2-3} 

                                                         & \cellcolor{white}-               &\cellcolor{lightgray!50}18. Vertical Place\\ \cline{2-3} 
                                                         & \cellcolor{white}-               &\cellcolor{darkgray!50}19. Pour
                                                         \\ \cline{2-3}        
                                                         & \cellcolor{white}-               &\cellcolor{teal!75}20. Release                     
                                                          \\ \hline
\end{tabular}
\label{tab:skills}
\end{table}

\begin{table}[t]
    \centering
    \caption{Long-Horizon tasks: primitive skill recombinations. 
Tasks are denoted by alphabetical letters ranging from A to T. Objects utilized for the demonstrations are indicated within parentheses: (s) sponge, (t) tomato passata package, and (b) bottle containing liquid. }
    \label{tab:lh_tasks}
    \begin{tabular}{|l|*{10}{>{\columncolor{white}}c|}}
        \hline
        \multicolumn{1}{|c|}{\textbf{Task}} & \multicolumn{10}{c|}{\textbf{Skills}} \\
        \hline
        & I & II & III & IV & V & VI & VII & VIII & IX & X \\
        \hline
         A (s)& \cellcolor{blue!25}1 & \cellcolor{yellow!75}5 & \cellcolor{red!25}3 & \cellcolor{orange!50}4 & 
         \cellcolor{magenta!50}7&
         \cellcolor{cyan!25}6 &  \cellcolor{violet!25}8 & \cellcolor{purple!50}9 & \cellcolor{teal!50}10 & \cellcolor{teal!75}20 \\
        \hline
         B (t)& \cellcolor{orange!50}4 & \cellcolor{magenta!50}7 & \cellcolor{violet!25}8 & \cellcolor{purple!50}9 & \cellcolor{teal!50}10 & \cellcolor{brown!50}11 & \cellcolor{lime!75}12 & \cellcolor{yellow!25}2 & & \\
        \hline
         C (b)& \cellcolor{olive!50}13 & \cellcolor{orange!75}14 & \cellcolor{teal!50}10 & \cellcolor{cyan!75}15 & \cellcolor{magenta!75}16 & \cellcolor{green!75}17 & \cellcolor{lightgray!50}18 & & & \\
        \hline
         D (s)& \cellcolor{cyan!25}6 & \cellcolor{magenta!50}7 & \cellcolor{cyan!25}6 & \cellcolor{magenta!50}7 & \cellcolor{cyan!25}6 & \cellcolor{magenta!50}7 & & & & \\
        \hline
         E (b)& \cellcolor{yellow!75}5 & \cellcolor{violet!25}8 & \cellcolor{purple!50}9 & \cellcolor{teal!50}10 & \cellcolor{cyan!75}15 & \cellcolor{darkgray!50}19 & & & & \\
        \hline
         F (b)& \cellcolor{violet!25}8 & \cellcolor{purple!50}9 & \cellcolor{teal!50}10 & \cellcolor{green!75}17 & & & & & & \\
        \hline
         G (b)& \cellcolor{blue!25}1 & \cellcolor{yellow!75}5 & \cellcolor{violet!25}8 & \cellcolor{purple!50}9 & & & & & & \\
        \hline
         H (t)& \cellcolor{cyan!75}15 & \cellcolor{magenta!75}16 & \cellcolor{cyan!75}15 & \cellcolor{lime!75}12 & & & & & & \\
        \hline
        I (s)& \cellcolor{magenta!75}16 & \cellcolor{cyan!75}15 & \cellcolor{magenta!75}16 & \cellcolor{teal!75}20 & & & & & & \\
        \hline
         J (b)& \cellcolor{purple!50}9 & \cellcolor{teal!50}10 & \cellcolor{green!75}17 & \cellcolor{teal!75}20 & & & & & & \\
        \hline
         K (t)& \cellcolor{orange!50}4 & \cellcolor{violet!25}8 & \cellcolor{purple!50}9 & & & & & & & \\
        \hline
         L (s)& \cellcolor{olive!50}13 & \cellcolor{orange!75}14 & \cellcolor{green!75}17 & & & & & & & \\
        \hline
         M (s)& \cellcolor{purple!50}9 & \cellcolor{teal!75}20 & \cellcolor{yellow!25}2 & & & & & & & \\
        \hline
         N (s)& \cellcolor{green!75}17 & \cellcolor{teal!50}10 & \cellcolor{magenta!75}16 & & & & & & & \\
        \hline
         O (b)& \cellcolor{teal!50}10 & \cellcolor{green!75}17 & \cellcolor{darkgray!50}19 & & & & & & & \\
        \hline
         P (t)& \cellcolor{darkgray!50}19 & \cellcolor{green!75}17 & \cellcolor{lightgray!50}18 & & & & & & & \\
        \hline
         Q (s)& \cellcolor{yellow!75}5 & \cellcolor{violet!25}8 & \cellcolor{yellow!25}2 & & & & & & & \\
        \hline
         R (b)& \cellcolor{blue!25}1 & \cellcolor{olive!50}13 & \cellcolor{yellow!25}2 & & & & & & & \\
        \hline
         S (s)& \cellcolor{lightgray!50}18 & \cellcolor{teal!50}10 & \cellcolor{teal!75}20 & & & & & & & \\
        \hline
         T (b)& \cellcolor{teal!50}10 & \cellcolor{green!75}17 & \cellcolor{lightgray!50}18 & & & & & & & \\
        \hline
    \end{tabular}
\end{table}












\subsection{Long-Horizon Tasks}\label{sec:LH TASK}
As the number of skills increases, the combinations grow exponentially. Although it is possible to have a sequence with many different skills that can turn into a challenging demonstration, even a basic task repeated multiple times may involve just three skills, . For instance, a sequence comprising "Wipe Forth" and "Wipe Back" combined with other skills like "Pick" and "Place" or "Twist" may resemble a dish-washing operation. Similarly, an experiment involving "Shake Down," "Shake Up," and "Pour" might evoke the actions of mixing and serving a cocktail.

For this study, we have identified 20 sparse long-horizon tasks of varying complexity, as outlined in Table \ref{tab:lh_tasks}, which define the experimental set for the classifier. In this table, each long-horizon task is represented by a numbered skill sequence and with the same colours according to  Table \ref{tab:skills}.
To illustrate the independence of skills from each other in skill recombination, it has been occasionally chosen to initiate from pre-established verified conditions, as observed in tasks B, D, J, K, and M.
We will evaluate the performance of our framework on these combinations in Section \ref{sec::Classifier_Evaluation}. 

In the experiments section, we will replicate autonomously long-horizon tasks defined by sequences A and B. As detailed in the section \ref{sec::LH_task}, this will be achieved from learning from the single skills demonstrations used for classifier training. Thus, no long-horizon sequences will be used in any of the training phases.

\section{Data Collection}
This section delineates the data collection methodology used to evaluate the proposed approach, while Section \ref{sec:Results} offers a discussion of the achieved results. For all the experiments, it is assumed that the position of the object is known. 

The data collection is organized in three phases:
\begin{enumerate}[label=\Alph*.]
\item The training data for the 20 primitive skills were gathered through human teleoperation of the robot, with the majority of skills documented across 10 demonstrations involving a single object. Specifically, for the skills "Touch," "Wipe Back," and "Grasp," data were collected using both a soft sponge (16x8.5x4.5cm, 20g) and cardboard packages (14x7.5x4 cm, 15g), with 20 demonstrations performed for each skill. Additionally, for the skill "Lift with Grasp," an extra 10 demonstrations were conducted using a bottle (22x7x7cm, 220g). This collected data was then used to train the model for the autonomous primitive skill classifier, as shown in Figure~\ref{Fig:fig2NNmodel}.
\item 
The first phase of experiments is made to evaluate the performance of the classifier. The demonstrations of the 20 long-horizon tasks presented in Table \ref{tab:lh_tasks} were collected. Each demonstration of a long-horizon task was presented and manually segmented with labels to make a comparison with the predictions obtained by the classifier. This process is depicted in Fig. \ref{Fig:Figure_1}. 
\item 
The second set of experiments consists of replacing the Leader agent with distinct MLP controllers shown in Fig. \ref{Fig:fig3Controlarch}, each tailored to a specific skill, trained from the same demonstrations used for the classifier. As in the methods section \ref{sec::LH_method}, on receiving a skill label, the robot reproduces the single learned skill and switches to the next one on receipt of a new skill label.
\end{enumerate}

\begin{figure}[t]
\centering
\includegraphics[width=1.0\linewidth, keepaspectratio]{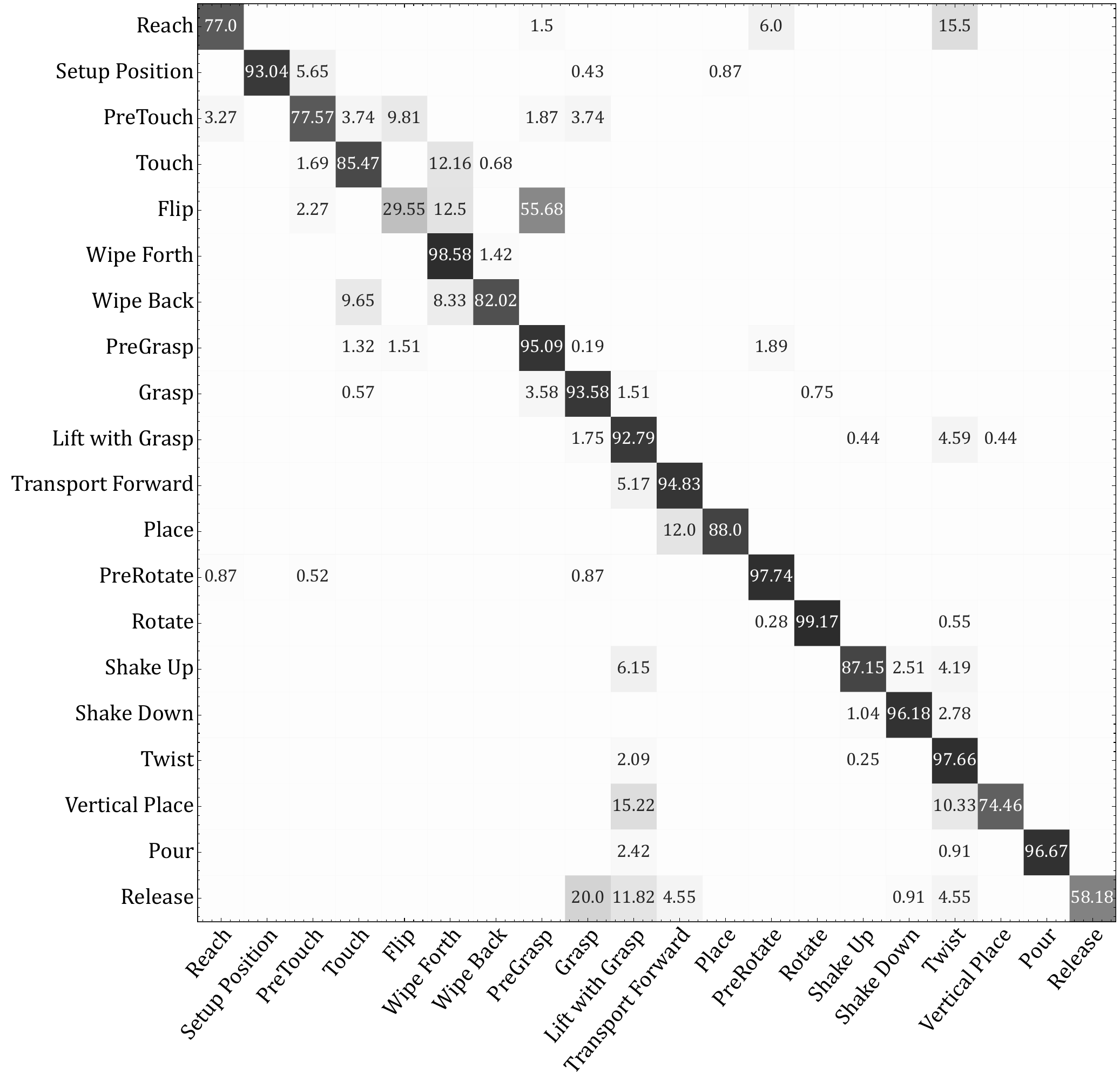}
\caption{Confusion matrix ($\%$) of the segmentation system on the Long-horizon demonstrations (detailed in Section~\ref{sec:LH TASK}).}
\label{confusion_matrix}
\vspace{-5mm}
\end{figure}

\section{Experimental Results}\label{sec:Results}
In this section, we present the evaluation of our framework through a series of experiments. We explore the impact of integrating an auto-regressive AE on the accuracy of skill label prediction. Additionally, we investigate the benefits of jointly training the auto-regressive AE with the label decoder for overall performance improvement. Our experiments also validate the efficiency of the proposed haptic features in accurately differentiating between primitive skills. Finally, we demonstrate the ability of our framework to efficiently select and integrate skill primitives for the execution of complex, long-horizon manipulation tasks.

\subsection{Classifier Evaluation}\label{sec::Classifier_Evaluation}
We train the classifier using data collected from primitive skills and validate its performance on unseen long-horizon manipulation tasks. To validate the effectiveness of our framework, we choose parameters including Accuracy, Precision, Recall, F1-score, and Average IoU as our key performance indicators.

\begin{figure}[t]
\centering
\includegraphics[width=1.0\linewidth,keepaspectratio]{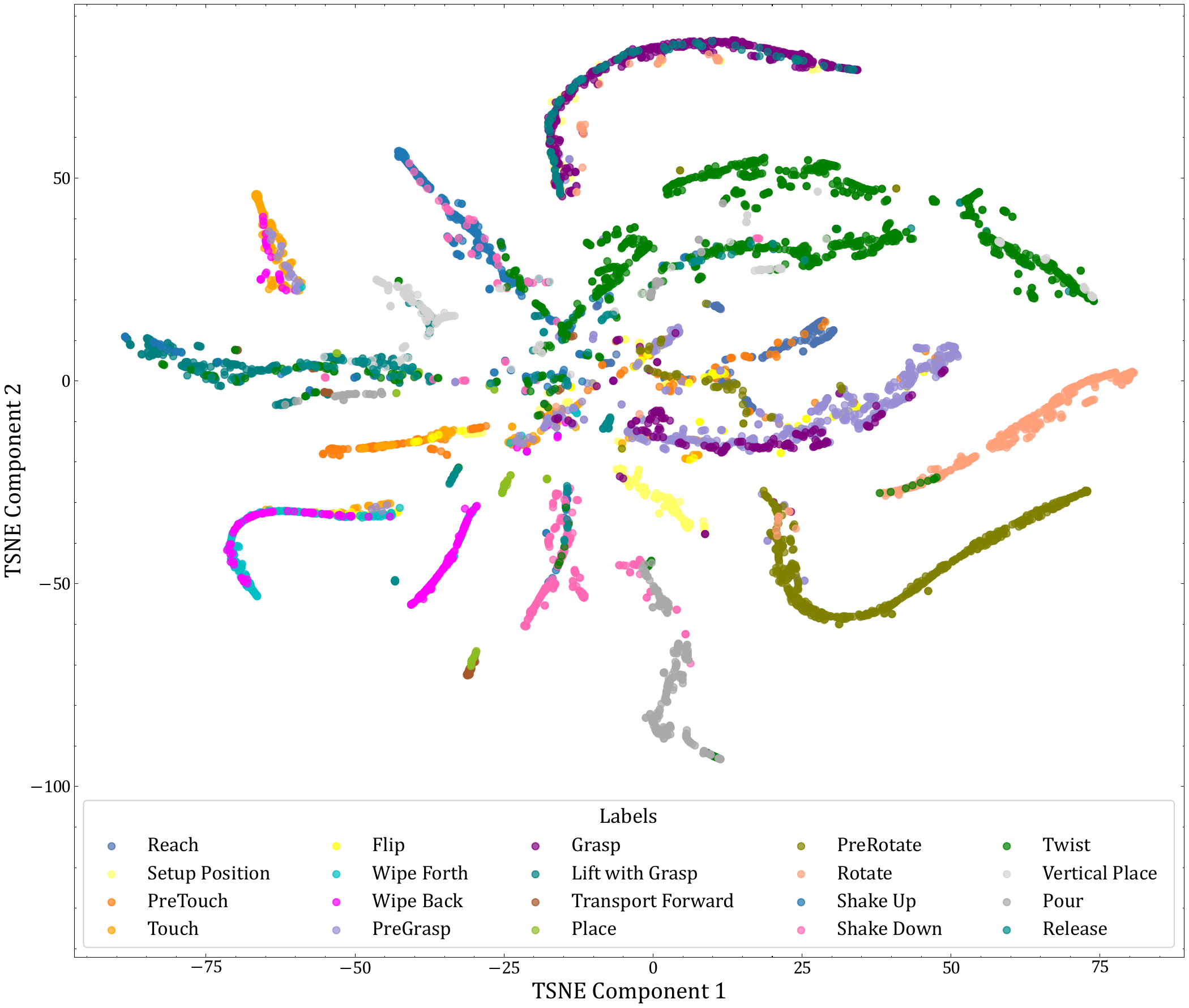}
\caption{
T-SNE visualization of the classifier latent features. Each point in the graph corresponds to a primitive skill instance, differentiated by various colours to distinguish among the primitive skills. 
}
\label{tsne_2d}
\vspace{-5mm}
\end{figure}

\subsubsection{Framework Validation}
We validate the performance of our proposed framework for long-horizon dexterous manipulation task segmentation by comparing it against three baselines. First, we examine a setup that utilizes a pre-trained AE for feature encoding, which is then followed by training a label decoder on these encoded features. Then, our second baseline explores the combination of a feature recovery AE with a label decoder, thereby excluding the temporal prediction from the AE.
Finally, we verified the effectiveness of training the feature encoder and the label decoder together, excluding the temporal feature decoder. The outcomes are presented in Table~\ref{tab:results}, where "LD" denote the label encoder. Our framework achieved an accuracy of 91\%, outperforming all baseline comparisons. This indicates that jointly training the temporal AE with the label decoder is beneficial for capturing the latent dynamics of robot behaviour while preserving essential information critical for differentiating between primitive skills. 
This capability significantly contributes to improving the performance of the classifier by ensuring a deeper comprehension of the temporal features and movements associated with various primitive skills.

\begin{table*}[]
    \centering    
    \caption{Comparative Performance Evaluation}
    \label{tab:results}
\begin{tabular}{l|llllll}
            & DexSkills & Pre-trained AE+ LD & Feature recovery AE + LD & Feature Encoder + LD & Raw haptic data & Calculated feature \\ \hline
Accuracy    & 0.91                    &   0.83      &    0.73      &         0.83              &      0.35       & 0.76               \\
Precision   & 0.89                    &    0.81     &     0.64     &           0.80             &       0.40      &  0.77             \\
Recall      & 0.85                    &    0.76     &     0.66     &            0.76        &       0.37        &   0.69          \\
F1-Score    & 0.87                    &    0.76     &     0.63     &          0.75               &      0.34       & 0.69              \\
Average IoU & 0.79                    &     0.64    &    0.56      &         0.63                &      0.23       &  0.55             
\end{tabular}
\end{table*}

\subsubsection{Feature importance verification}
To verify the importance of the proposed feature for skill classification, we evaluated the performance of our proposed framework in two distinct scenarios:
firstly, the framework was trained on raw haptic data, including the end-effector state, filtered tactile information, filtered contact indicators, and the AH joint state, which are the data used in training the policy for primitive skills; secondly, by training only with the proposed features while excluding the raw haptic data. The results are shown in Table~\ref{tab:results}. When trained with raw haptic information, the classifier achieves an accuracy of only 35\%. In contrast, utilizing the additional features proposed for skill segmentation improves the accuracy to approximately 76\%. This substantial improvement indicates the effectiveness of our proposed features for skill segmentation, highlighting their value in enhancing the performance of the classifier.

\subsubsection{Confusion matrix for each skill}
The confusion matrix provides a comprehensive overview of how well a classification model is performing across all categories. The result of the confusion matrix on 20 unseen long-horizon tasks is shown in Fig.~\ref{confusion_matrix}. 
The prediction outcomes for the majority of the skills are satisfactory.
However, the skills of flip and release tend to be confused with other skills, primarily due to their short duration, which makes their features challenging to capture accurately. Additionally, it shall be noted that manual labelling of the ground truth in long-horizon tasks can introduce certain errors inevitably, which can become especially obvious during the transition phases between skills.

\begin{figure}[t]
\centering
\begin{subfigure}{\linewidth}
  \centering
  \includegraphics[width=1.0\linewidth, keepaspectratio]{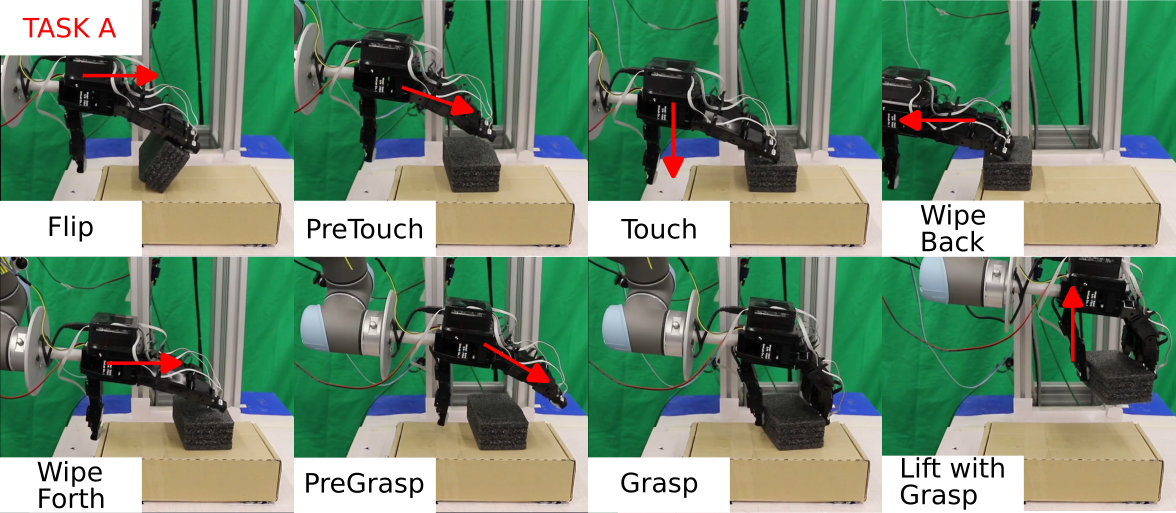}
  \caption{Long-horizon Task A using a soft sponge (16x8.5x4.5cm, 20g).}
  \label{fig:video_A}
\end{subfigure}
\par\bigskip
\begin{subfigure}{\linewidth}
  \centering
  \includegraphics[width=1.0\linewidth, keepaspectratio]{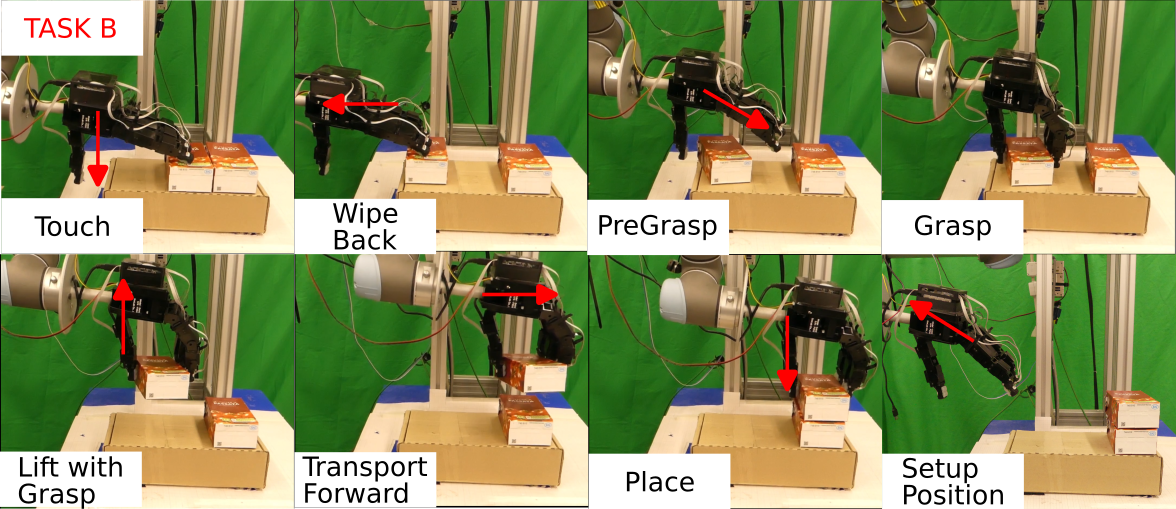}
  \caption{Long-horizon Task B using a cardboard package (14x7.5x4 cm, 15g).}
  \label{fig:video_B}
\end{subfigure}
\caption{Autonomous skills reproduction. Full videos are available on the project repository}
\label{fig:combined}
\vspace{-5mm}
\end{figure}

\subsection{Autonomous Long-horizon Tasks Execution}\label{sec::LH_task}

In this work, we verified our framework on two challenge tasks, which are identified as tasks A and B in Table~\ref{tab:lh_tasks}. Each task required the coordination of over 8 distinct skills, with skill transitions manually controlled by a human operator.  Both tasks were completed by using the predicted primitive skills combination sequence. This demonstrates the ability of our proposed framework to reuse primitive skills for achieving unseen long-horizon tasks effectively. The snapshots of these experiments are presented in Figures~\ref{fig:video_A} and~\ref{fig:video_B}. Additionally, videos detailing the experimental process are available on the project GitHub page\footnotemark[\value{footnote}].

\subsection{T-SNE Analysis}
We employed t-Distributed Stochastic Neighbor Embedding (t-SNE) analysis on the latent features extracted before the final layer of the label decoder. This technique projects high-dimensional vectors onto a two-dimensional plane, achieving the visualization of high-dimensional data relationships. The application of t-SNE to our unseen long-horizon task data is depicted in Fig.~\ref{tsne_2d}, where a distinct colour represents each skill. 

Overall, the t-SNE plot demonstrates that most skills are well-clustered and exhibit clear distinctions from one another. However, for skills that exhibit similar behaviours or tend to occur sequentially such as "PreGrasp", "Grasp", and "Lift with Grasp", there is a noticeable overlap in their representation. This overlapping tendency is also observed among skills with similar motions, like "Wipe Back" and "Wipe Forth". 

\section{Conclusion}
In this study, we have presented DexSkills a learning framework that addresses the challenge of executing real-world long-horizon tasks with dexterous robotic hands by decomposing them into reusable primitive skills, trained from human demonstrations. This framework, adaptable for various demonstrations such as bilateral or kinesthetic teleoperation, enables the segmentation of long-horizon demonstrations into individual tasks, using only proprioceptive and tactile data, facilitating their integration into the repertoire of known skills and enabling autonomous robot execution of diverse tasks with high classification accuracy.
In this study, we evaluate the segmentation performance of unseen multiple long-horizon tasks starting from a limited set of primitive skills. Our framework demonstrated strong performance in skill segmentation, achieving an accuracy rate of $91\%$. Comparative analysis reveals that our framework surpasses alternative approaches by effectively capturing the latent dynamics of primitive skills within the feature set. 
Furthermore, through the same demonstrations employed for the Classifier, a protocol is presented for learning the autonomous control over each individual task. Although the accuracy was not studied, it is established how each of the demonstrated tasks can be replicated autonomously by the robot by providing a label sequence. Future work should include the development of more complex models for autonomous reproduction and should accurately assess the reproducibility of autonomous execution of long-horizon tasks.

\bibliographystyle{IEEEtran}
\bibliography{references}

\end{document}